\documentclass{article}
\usepackage{ijcai22}

\usepackage{times}
\usepackage{soul}
\usepackage{url}
\usepackage[hidelinks]{hyperref}
\usepackage[utf8]{inputenc}
\usepackage[small]{caption}
\usepackage{graphicx}
\usepackage{amsmath}
\usepackage{amsthm}
\usepackage{booktabs}
\usepackage{algorithm}
\usepackage{algorithmic}
\usepackage{hyperref}

\usepackage{enumerate}
\usepackage[aboveskip=1pt]{subcaption}
\usepackage{xcolor}

\title{Dynamic Domain Generalization}

\author{
Zhishu Sun$^1$\and
Zhifeng Shen$^1$\and
Luojun Lin$^{1,}$\footnotemark[2]\and
Yuanlong Yu$^{1,}$\footnotemark[2]\and
Zhifeng Yang$^1$ \\
Shicai Yang$^{2}$\And
Weijie Chen$^2$ \\
\affiliations
$^1$College of Computer and Data Science, Fuzhou University, Fuzhou, China\\
$^2$Hikvision Research Institute, Hangzhou, China\\
\emails
linluojun2009@126.com, 
yu.yuanlong@fzu.edu.cn
}
\begin{document}

\maketitle
\renewcommand{\thefootnote}{\fnsymbol{footnote}}
\footnotetext[2]{Corresponding authors: Luojun Lin and Yuanlong Yu}
\renewcommand{\thefootnote}{\arabic{footnote}}

\begin{abstract}
    %   Generalizing to unseen domains is a challenging topic with practical research significance for advanced vision tasks. Most of existing researches involve using domain labels, e.g., domain alignment or augmentation through adversarial training. However, these methods are ill-suited for most real scenarios where domain labels are ambiguous or inaccessible, e.g., the web-crawled data. To relax this issue, we propose an domain-label-free method based on \textbf{dynamic convolution}, where the kernel parameters are adapted with instances. Specifically, the dynamic convolution is formulated by aggregating static convolution kernels and dynamic convolution kernels, where the dynamic kernels are designed with asymmetric convolution for the sake of strengthening kernel skeletons. The key insight of our method is to achieve instance-aware domain generalization that learns feature alignment in instance-level, which is more fine-grained than domain-level feature alignment methods. Moreover, in inference stage, the model still can be adapted to agnostic target samples automatically, which is difficult to be achieved by static convolution in conventional CNN architectures. Experimental results show that, without using domain labels, our dynamic convolution is superior to other CNN architectures and domain-label-free methods, moreover, it can be flexibly combined with other methods to achieve state-of-the-art results (\textbf{90.18\%} accuracy on PACS and \textbf{46.93\%} accuracy on DomainNet). 
    Domain generalization (DG) is a fundamental yet very challenging research topic in machine learning. The existing arts mainly focus on learning domain-invariant features with limited source domains in a static model. Unfortunately, there is a lack of training-free mechanism to adjust the model when generalized to the agnostic target domains. To tackle this problem, we develop a brand-new DG variant, namely \emph{Dynamic Domain Generalization} (DDG), in which the model learns to twist the network parameters to adapt the data from different domains. Specifically, we leverage a meta-adjuster to twist the network parameters based on the static model with respect to different data from different domains. In this way, the static model is optimized to learn domain-shared features, while the meta-adjuster is designed to learn domain-specific features. To enable this process, DomainMix is exploited to simulate data from diverse domains during teaching the meta-adjuster to adapt the upcoming agnostic target domains. This learning mechanism urges the model to generalize to different agnostic target domains via adjusting the model without training. Extensive experiments demonstrate the effectiveness of our proposed method. {Code is available: {\color{cyan}\url{https://github.com/MetaVisionLab/DDG}}}
    % \footnote{Code is available at: {\color{cyan}\url{https://github.com/MetaVisionLab/DDG}}}
\end{abstract}

\section{Introduction}

\begin{figure}[t]
     \centering
     \footnotesize
     \vspace{-0.2cm}
     \includegraphics[width=0.8\columnwidth]{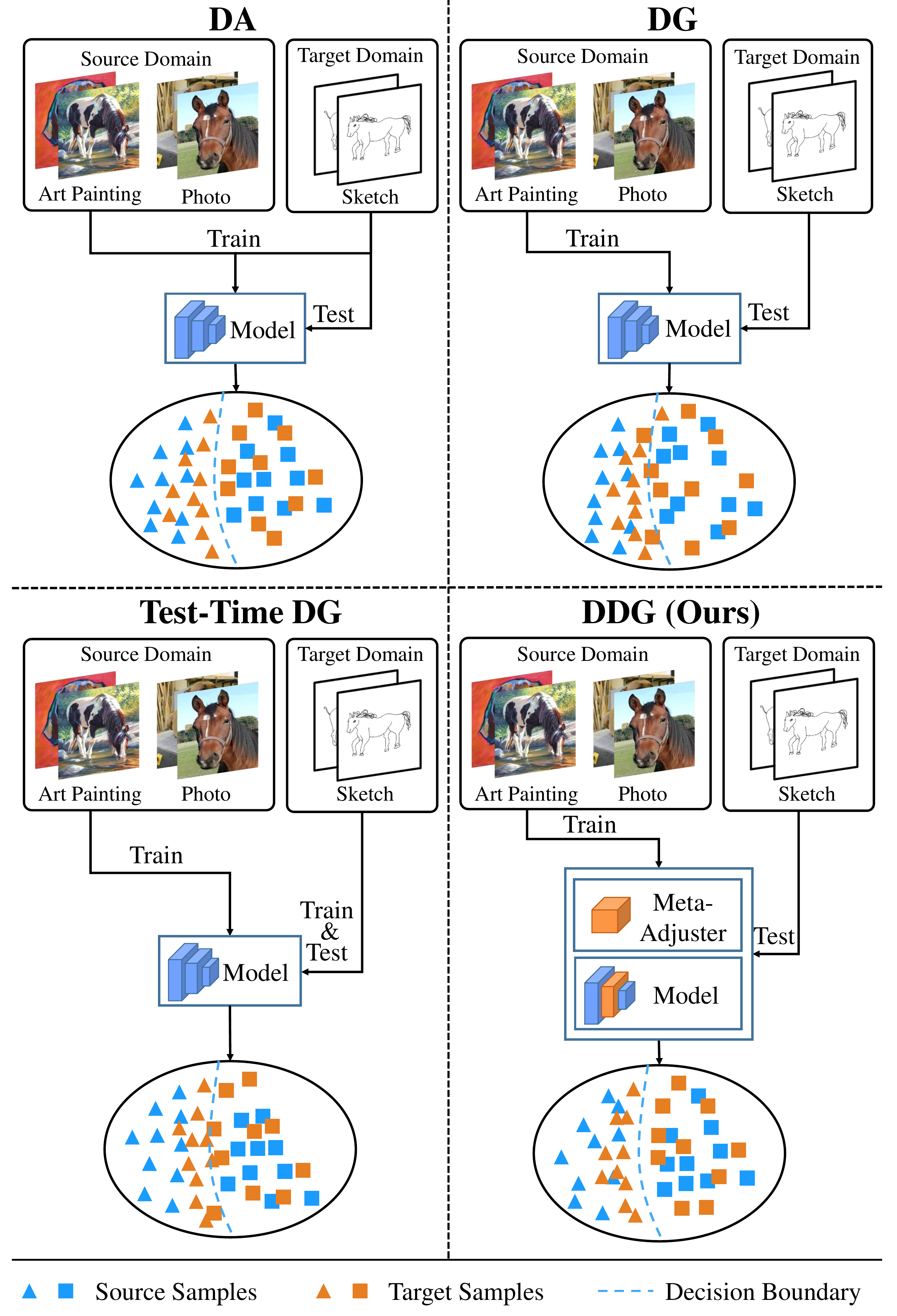}
     \vspace{-0.2cm}
     \caption{Comparison among DA, DG, Test-time DG and DDG methods. DA and test-time DG are able to adjust the model when encountering novel target domains via fine-tuning, while DG freezes a static model to adapt any agnostic target domain. Different from the existing methods, DDG is able to adjust the model without any training effort for better domain generalization.}
     \label{fig:ddg}
     \vspace{-0.2cm}
\end{figure}

Deep neural networks have achieved great success on various vision tasks, such as image recognition and object detection~\cite{resnet}. Most of deep models are learnt in a closed perception environment where training and testing data are supposed under \emph{i.i.d} (independent and identical distribution) assumption. However, the reality is open, compound, uncontrolled and incompatible with \emph{i.i.d} assumption. Such complex scenarios can make conventional models suffer tremendous performance degradation, due to the absence of generalization ability to handle domain shift, which undoubtedly will discount the reliability and security of intelligent systems, \emph{e.g.}, autonomous driving system.

Extensive studies aim to tackle this problem through domain generalization (DG), whose objective is to obtain a robust static model by learning domain-invariant representations via exploiting multiple source domains~\cite{dann,mixstyle}. Nevertheless, a severe bottleneck from the static model constrains its freedom to generalize to agnostic target domains with different distribution. In contrast, domain adaptation (DA)~\cite{ganin2016domain} and test-time DG~\cite{iwasawa2021test} are explored to leverage the unlabeled data sampled from the target domain for further unsupervised fine-tuning so as to enhance the target-oriented generalization ability. However, extra training effort and latency are required during testing, while the manufactured chips and edge devices mostly do not possess training resources. Considering these two perspectives, there is a lack of training-free mechanism to adjust the model when encountering the testing data from novel target domains. 

To complement this blank, we propose a new DG variant, \emph{Dynamic Domain Generalization} (DDG). As shown in Figure \ref{fig:ddg}, different from DA, DG, as well as test-time DG methods, the proposed DDG is attached with a meta-adjuster, which learns to adapt novel domains during pre-training and twists the model during testing without any fine-tuning. To drive the process of learning to adapt, we need to collect training data from numerous novel domains, while the reality is that we are usually hindered by limited source domains. A simple yet friendly solution is that we simulate novel samples from the limited source domains via a random convex interpolation among different source domains, which is termed as DomainMix in this paper. In this way, each simulated sample is restricted to a unique domain. As a consequence, the domain-aware meta-adjuster can be relaxed into an instance-aware meta-adjuster, which modulates the network parameters per instance. Besides, it makes the model easier adapt from source domains to agnostic target domains with undefinable domain distinction, by discarding the utilization of domain label during optimizing meta-adjuster.

Beyond the optimization pipeline of DDG, it is also a critical factor on how to design the meta-adjuster. In order to disentangle the domain-shared features and domain-specific features, the network parameters are decoupled into a static part and a dynamic part where the latter one is generated by the meta-adjuster per instance. To ease the optimization of meta-adjuster, the dynamic parameters are decomposed into a linear combination of several elaborately-designed kernel templates with low-dimensional dynamic coefficients, in which the coefficients are generated dynamically by the meta-adjuster via taking the instance embedding as input. As shown in Figure \ref{fig:block}, the kernel templates are designed as four asymmetric kernels for the sake of strengthening the kernel skeletons in spatial as well as channel dimensions, which explicitly regularize the meta-adjuster to generate diverse dynamic parameters to adapt a large variety of novel samples.

Extensive experiments are conducted on three popular DG benchmarks, including PACS \cite{pacs}, Office-Home \cite{venkateswara2017OfficeHome} and DomainNet \cite{DomainNet}. These datasets contain various situations varying from small to large scales, from simple to complex scenarios, from small to large domain shifts, which are sufficient to demonstrate the effectiveness of our proposed method. To summarize, DDG opens up a new paradigm to study DG problem.

\section{Related Work}
\paragraph{Domain Adaptation.}
Conventional DA aims to transfer the knowledge learnt from labeled source domains to target domains which may contain a few labeled samples or none at all. 
The discrepancy-based methods are early developed, which mitigate the distribution discrepancy between two domains by minimizing the well-defined distance loss \cite{haeusser2017associative}. Another line of DA methods is based on adversarial learning, where a domain discriminator is frequently-used to train with the feature extractor adversarially, in order to enforce the feature extractor to learn domain-invariant features by confusing the domain discriminator~\cite{ganin2016domain}. Compared with DA methods that adjust models to adapt target domains explicitly, our method can be viewed as a kind of training-free DA via pre-training merely in source domains.

\paragraph{Domain Generalization.}
Early DG-related research focus on learning domain-invariant representation, which is believed to be robust if it can suppress domain shift among multiple source domains~\cite{motiian2017unifiedDG}. Another line of DG methods is based on data augmentation that enhances the diversity of image styles for optimizing the generalization capability of models~\cite{ddaig,mixstyle}. Self-supervised learning is also applied to DG that learns task-independent representation with less probability of over-fitting~\cite{carlucci2019JiGen,kim2021selfreg}. Meta-learning is also involved with solving DG~\cite{balaji2018metareg}, where deep model is trained by simulating the meta-train and meta-test tasks in source domains to enhance its generalization ability. Our method can be regarded as a very simple meta-learning-like paradigm, in which meta-adjuster learns to learn parameter generalization for each individual ``domain''. 
Besides, a few emerging works show a new insight for DG that deep model can be adapted to target domain at test time~\cite{iwasawa2021test}. However, the computational cost is inescapable in such methods. In contrast, our method can directly update parameters to adapt agnostic target domain without any backward propagation.

\begin{figure}[t]
     \centering
     \footnotesize
     \includegraphics[width=1\columnwidth]{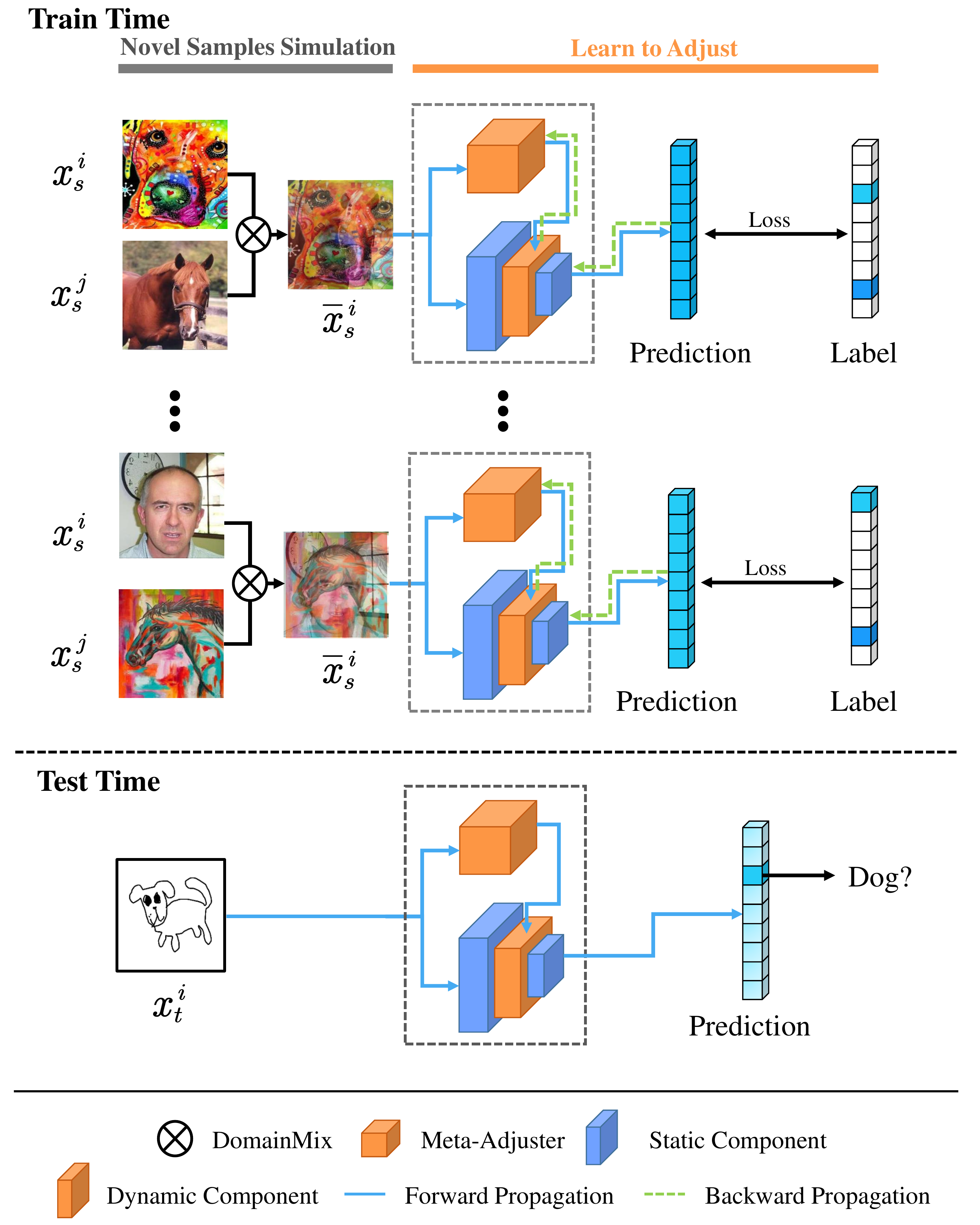}
     \vspace{-0.2cm}
     \caption{Pipeline of the proposed DDG method. In the training phase, two stages are required, namely novel samples simulation and learning to adjust, which aims to teach the meta-adjuster to adapt novel samples. In the testing phase, the network can be adjusted without training when encountering agnostic target domain.}
     \label{fig:pipeline}
     \vspace{-0.2cm}
\end{figure}

\section{Methods}

\subsection{Overview}
DG aims to train a general model that can be well generalized to any agnostic target domain. In particular, we only consider $K$-way classification task in this paper. For a vanilla DG task, we are given $n_s$ labeled samples $\{x_s^i, y_s^i, d_s^i\}_{i=1}^{n_s}$ drawn from multi-source domains $\mathcal{X}_S=\{\mathcal{X}_{S_1}, \mathcal{X}_{S_2},..., \mathcal{X}_{S_M}\}$, where $x_s^i \in \mathcal{X}_S$, $y_s^i \in \mathcal{Y}_S$ and domain label $d_s^i\in [1,M]$. The objective of DG is to learn a mapping $f_\Theta: \{\mathcal{X}_{S_1} ,..., \mathcal{X}_{S_M}\}\rightarrow \mathcal{Y}_S$ that can also predict precisely on any agnostic target domain. 

\paragraph{Vanilla Domain Generalization.}
For vanilla DG, the learnt mapping $f_\Theta$ obtained in training stage will be kept static in inference. It usually causes the performance degradation on agnostic target domains due to the inevitable data distribution distinction between the source and target domains. 

\paragraph{Dynamic Domain Generalization.}
To overcome the weakness of vanilla DG, we propose a new variant of DG, namely \emph{Dynamic Domain Generalization} (DDG). DDG mainly focuses on learning to twist the network parameters to adapt the instances from diverse novel domains. In this paper, each instance is assumed to be an independent latent domain. Under this assumption, the model parameters $\Theta$ can be modeled as a function of input instance $x$: $\Theta=\Theta(x)$. To this end, a meta-adjuster is designed to generate network parameters to achieve diverse networks so as to adapt diverse novel samples.
In this case, the mapping $f_{\Theta(x)}$ is more likely to align different instances into the same feature space under the reflection of instance-aware parameters $\Theta(x)$. Compared with the static mapping that is difficult to align instances from different data distributions, DDG is easier to achieve feature alignment among diverse novel domains. 

The pipeline of our method is illustrated in Figure \ref{fig:pipeline}, where DDG framework is composed of a static component and a dynamic component updated by the meta-adjuster based on each input instance. The former one is responsible to learn domain-shared features while the latter one is designed to learn domain-specific features. In order to teach the meta-adjuster to adapt diverse novel domains, we simulate numerous novel training samples via a random convex interpolation among the limited source domains. 

% The more diverse domains will benefits model to enforce the feature to fill in the entire space, which will make the model more robust. 
% Further, we implement a DomainMix method that mixes the instances from different domains to obtain a large amount of synthetic instances with diverse stylized domains. In training stage, the more diverse inputs will drive meta-adjuster to generate more varied parameters to enforce the feature to fill in the entire space, which will make the model more robust.   

\subsection{Novel Samples Simulation}
Numerous novel samples are required to drive the optimization of learning to adjust. By exploiting the limited source domains, we develop a DomainMix method from MixUp~\cite{mixup} to synthesize novel domains with diverse styles. Given two instances $(x_s^i, y_s^i)$ and $(x_s^j, y_s^j)$ randomly sampled from two different source domains, the process of novel samples simulation is formulated as:
\begin{equation}
\bar{x}_s^i = \alpha x_s^i + (1-\alpha) x_s^j, \quad \quad \bar{y}_s^i = \alpha y_s^i + (1-\alpha) y_s^j,
\label{eq:domainmix}
\end{equation}
where $\alpha$ is randomly sampled from \emph{Beta} distribution in each feed-forward. Each DomainMix sample $(\bar{x}_s^i, \bar{y}_s^i)$ can be viewed as an unique domain and be harnessed to drive the following process of learning to adjust.

\subsection{Learn to Adjust}
\begin{figure}[t]
\centering
\includegraphics[width=\columnwidth]{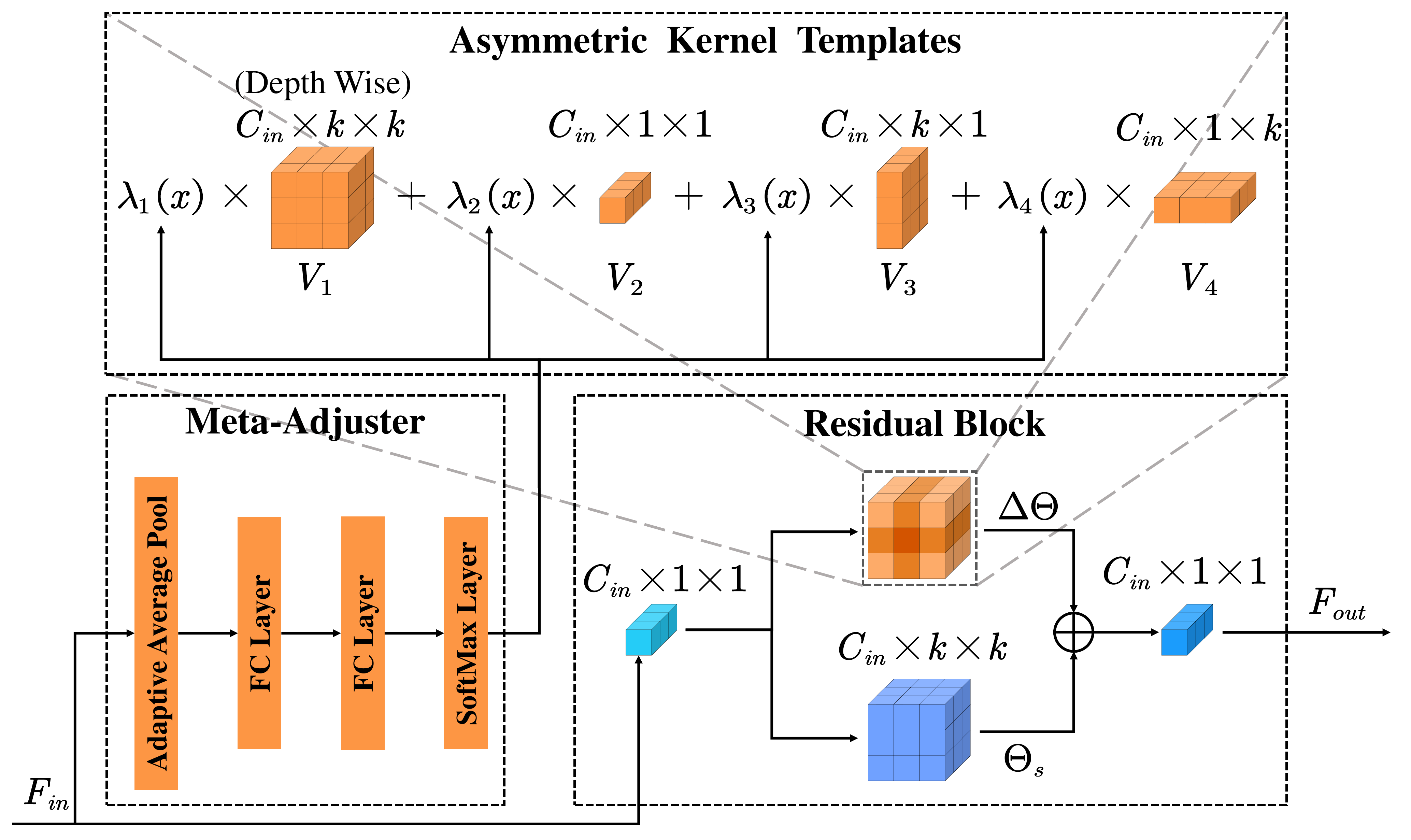}
\caption{Structure illustration of the basic block for DDG, which contains a static and a dynamic component. The latter one is a learnable linear combination of four asymmetric kernel templates and corresponding dynamic coefficients which are generated by the meta-adjuster. For simplicity, $C_{out}$ is omitted.}
\label{fig:block}
\end{figure}

In this section, we decouple the network parameters into a static part and a dynamic part which implicitly disentangles the domain-shared and domain-specific features. Here the dynamic parameters are generated by a shallow generator, termed as meta-adjuster, by taking the instance embedding as input. For a better illustration, we take one of the most popular convolutional neural networks, namely ResNet~\cite{resnet}, as the backbone of our dynamic network. As shown in Figure \ref{fig:block}, the meta-adjuster is embedded in each middle convolution layer in the residual block (note that each residual block is composed by a sequential of $1 \times 1$, $3 \times 3$ and $1 \times 1$ convolution layers). In the following, we will introduce how to design meta-adjuster effectively.

% It aims to enforce the dynamic parameters distributed in a small subset, which will benefit optimization. 

% In the following, the details of dynamic mapping will be introduced, by taking residual network~\cite{resnet} as example, where the dynamic mapping module is implemented in each middle convolution layer of residual block, as shown in Figure  \ref{fig:block}. 
% the residual block takes the output from last block as its input, which is denoted as $F_{in}$, and the meta-adjuster also take it as input. 

\paragraph{Meta-Adjuster.}
% Meta-adjuster generates residual kernels to add to the static kernels of the dynamic convolution layer, with the purpose of adjusting the kernels towards the instance-aware direction. 
Meta-adjuster takes the instance embedding $F_{in}(x)$ (\emph{i.e.}, the output of last block) as input, so that the meta-adjuster practically can be considered as the function of input instance $\boldsymbol{\Delta\Theta}(x)$. Therefore, the network parameters adjusting process can be formulated as:
\begin{equation}
\boldsymbol{\Theta}(x) = \boldsymbol{\Theta}_s +  \boldsymbol{\Delta\Theta}(x),
\label{eq:meta-adjuster}
\end{equation}
where $\boldsymbol{\Theta}_s$ and $\boldsymbol{\Theta}(x)$ represent the static parameters and its corresponding up-to-date parameters, and $\boldsymbol{\Delta \Theta}(x)$ denotes their dynamic term. Theoretically, the dynamic term $\boldsymbol{\Delta \Theta}(x)$ should be the same dimension with the static parameters $\boldsymbol{\Theta}_s$. However, it makes the amount of network parameters grow quadratically and then apparently increases the risk of over-fitting. Thus, according to matrix factorization, we propose to decompose the dynamic term into a learnable linear combination of several kernel templates $\{\boldsymbol{V}_n\}_{n=1}^N$ which can be formulated as:
\begin{equation}
\boldsymbol{\Delta\Theta}(x) = \sum_{n=1}^N \lambda_n(x) \boldsymbol{V}_n,
\label{eq:theta}
\end{equation}
where $N$ is a hyper-parameter that represents the number of kernel templates. $\{\lambda_n(x)\}_{n=1}^N$ denotes the set of dynamic coefficients, which is a group of instance-aware scalars generated by a learnable shallow meta-adjuster ${\lambda(\cdot)}$ via taking instance $x$ (practically the instance embedding $F_{in}(x)$) as input. In this way, the dynamic component are restricted into a limited space, which benefits the optimization of whole model. 

\paragraph{Asymmetric Kernel Templates.}
An apparent side effect to decompose meta-adjuster into a linear combination of kernel templates and a few dynamic coefficients is that the modulation brought by dynamic parameters are prone to be tiny and rigid. To avoid this tendency, we design four asymmetric kernel templates to explore the relationship between dynamic parameters from different dimensions, as well as to enforce the meta-adjuster to generate more diverse dynamic parameters for a large variety of novel samples. Another discovery is that the learned knowledge of convolutional kernels is not distributed uniformly, since the skeleton of kernels (\emph{i.e.} the central criss-cross positions) usually has larger weight values~\cite{ding2019acnet}. For this reason, we propose to enhance the skeleton of kernels by generating skeleton-based dynamic parameters that is derived from a linear combination of these asymmetric skeleton-based kernel templates.

Suppose that the dimension of static kernels $\boldsymbol{\Theta}_s$ is denoted as $c_{in}\times c_{out}\times k\times k$, where $k$, $c_{in}$ and $c_{out}$ represent the kernel size, the input channel number as well as the output channel number. As shown in Figure \ref{fig:block}, the asymmetric skeleton-based kernel templates are designed as four matrices with different shapes as $c_{in}\times 1 \times k \times k$, $c_{in}\times c_{out} \times 1 \times 1$, $c_{in}\times c_{out} \times k \times 1$ and $c_{in}\times c_{out} \times 1 \times k$, respectively. Note that the kernel template with size of $c_{in}\times 1 \times k \times k$ are implemented as depth-wise convolution, because it merely focuses on spatial relationship. In this way, the dynamic parameters are encouraged to be varied for different input samples. In addition, the skeleton of network parameters can be amplified in spatial as well as channel dimensions.
% which benefits to learn domain-shared knowledge.

% \paragraph{Learning to Adjust}
% \begin{equation}
% \begin{aligned}
% \mathcal{L}_{ce}(\mathcal{X}_S, \mathcal{Y}_S | \boldsymbol{\Theta}_s)
% \label{eq:mapping}
% \end{aligned}
% \end{equation}

% f_{\boldsymbol{\Theta}(x)}(x) &= f_{\boldsymbol{\Theta}_s+\boldsymbol{\Delta\Theta}(x)}(x)\\

% \paragraph{Dynamic Mapping}
% % The dynamic network is constituted by stacking the dynamic residual blocks in Figure  \ref{fig:block} to acquire dynamic mapping. 
% Due to the linear property of convolution, the dynamic mapping of the whole block (like Figure  \ref{fig:block}) can be decoupled as the static component and dynamic component, which can be formulated as: 
% \begin{equation}
% \begin{aligned}
% f_{\boldsymbol{\Theta}(x)}(x) &= f_{\boldsymbol{\Theta}_s}(x) +  \Delta f_{\boldsymbol{\Theta}(x)}(x)\\
% & = f_{\boldsymbol{\Theta}_s}(x) + \sum_{i=1}^I \lambda_i(x) f_{\boldsymbol{V}_i}(x).
% \label{eq:mapping}
% \end{aligned}
% \end{equation}
% In practically, Eq. \ref{eq:mapping} can be implemented by aggregating the outputs of $I+1$ parallel convolution layers with asymmetric kernel sizes, which is easy and convenient to conduct.

% \paragraph{Discussion}
% The static component learn domain-shared features, and the dynamic component learn domain-specific features. 

\section{Experiments}
% We conduct extensive experiments on several standard DG benchmark datasets which varies from small-scale and large-scale, small domain shift and large domain shift. The experimental results verify the effectiveness of our method by comparing with other DG methods. Besides, by combing with other DG methods, our DDG method achieves state-of-the-art results, which stresses our method is plug and play and flexible with other methods.

\subsection{Datasets and Experimental Setup} 
% \paragraph{Datasets}
\paragraph{Datasets.}
1) \textbf{PACS}~\cite{pacs} contains more than $9k$ images with seven categories collected from four domains, which includes: \textit{Photo}, \textit{Sketch}, \textit{Cartoon} and \textit{Art Painting}. 
2) \textbf{Office-Home}~\cite{venkateswara2017OfficeHome} contains $15,500$ images with $65$ categories with a few images per category. All images are sampled from four domains, including: \textit{Art}, \textit{Clipart}, \textit{Product} and \textit{Real-World}. 
3) \textbf{DomainNet}~\cite{DomainNet} is a large-scale benchmark containing $600k$ images with $345$ categories. The images are collected from six different domains which are \textit{Clipart}, \textit{Inforgraph}, \textit{Painting}, \textit{Quickdraw}, \textit{Real} and \textit{Sketch}.

\paragraph{Experimental Setup.}
For each dataset, we conduct leave-one-out strategy that randomly selects one domain for evaluation and leaves the remaining ones as source domains for training. Each experiment is repeated three times with different random seeds, and the mean and variance of these results are reported in this paper.
 
\subsection{Implementation Details}
All the experiments are conducted on RTX 3090 GPU with PyTorch 1.10.0. We use ResNet-50 as the backbone of our dynamic network, by attaching the meta-adjusters to the original residual blocks (as shown in Figure \ref{fig:block}), where the spatial size of kernel templates is set as $k=3$. Note that in each block, the meta-adjuster is implemented by a sequential of a global average pooling layer, a ReLU layer sandwiched between two fully-connected layers, and a SoftMax layer to output instance-aware coefficients. 

Before training DG tasks, we pre-train our network on ImageNet so as to compare with the existing works fairly. Then, we train DG tasks by fine-tuning the pre-trained model with SGD. 
The training objective is a cross-entropy loss function, exactly the same with vanilla DG. The asymmetric kernel templates and meta-adjuster are optimized with the main backbone simultaneously during training.
For PACS and Office-Home, the network optimization is set with batch size of 64, training epochs of 50, and the initial learning rate of 1e-3 decayed by cosine scheduler. While training on DomainNet, most of the hyper-parameters keep the same with that of PACS, except that the initial learning rate and max epoch are 2e-3 and $15$, and the mini-batches are fetched with random domain sampler strategy~\cite{mixstyle}, in order to ensure that each domain is uniformly sampled.

\subsection{Comparisons with the State-of-the-Art}

\begin{table*}[t]
  \centering
  \setlength\tabcolsep{12pt}
  \resizebox{0.9\textwidth}{!}{

     % Table generated by Excel2LaTeX from sheet 'PACS-resnet50'
    \begin{tabular}{c|ccccc}
    \toprule
    Method & Art Painting   & Cartoon & Photo & Sketch & Avg. \\
    \midrule
    D-SAM~\cite{d-sam} & 77.33  & 72.43  & 95.30  & 77.83  & 80.72  \\
    CSD~\cite{piratla2020efficient_csd} & 78.90±1.10 & 75.80±1.00 & 94.10±0.20 & 76.70±1.20 & 81.40  \\
    Epi-FCR~\cite{li2019epi-fcr} & 82.10  & 77.00  & 93.90  & 73.00  & 81.50  \\
    MASF \cite{dou2019masf} & 82.89±0.16 & {80.49±0.21} & 95.01±0.10 & 72.29±0.15 & 82.67  \\
    DDAIG \cite{ddaig} & 84.20±0.30 & 78.10±0.60 & 95.30±0.40 & 74.70±0.80 & 83.10  \\
    MetaReg \cite{balaji2018metareg} & {87.20±0.13} & 79.20±0.27 & 97.60±0.31 & 70.30±0.18 & 83.60  \\
    MixStyle \cite{mixstyle} & 84.10±0.40 & 78.80±0.40 & 96.10±0.30 & 75.90±0.90 & 83.70    \\
    ERM \cite{gulrajani2020search_erm} & \underline{88.10±0.10} & 77.90±1.30 & \textbf{97.80±0.00} & 79.10±0.90 & 85.70  \\
    RSC \cite{huang2020rsc}&  87.89  &  82.16  & 83.35  &   97.92 & 87.83 \\
    SWAD \cite{cha2021swad}&  \textbf{89.30±0.20}  &  \textbf{83.40±0.60}  & 97.30±0.30  &   82.50±0.50 & \textbf{88.10} \\
    \midrule
    Ours  & 85.92±0.17 & 79.68±0.42 & 96.65±0.15 & \underline{82.62±0.27} & {86.21} \\
    Ours w/ MixStyle & 87.14±0.21 & \underline{82.66±0.33} & \underline{97.77±0.06} & \textbf{83.89±0.14} & \underline{87.87} \\
    \bottomrule
    \end{tabular}
  }
  \vspace{-0.2cm}
  \caption{Results on PACS dataset. The best and second-best results are \textbf{bold} and \underline{underlined}, respectively.} %$^*$: The results are taken from the original literature.}% $\ddagger$: The results are re-implemented by ours.}
 \vspace{-0.2cm}
  \label{tab:pacs}%
\end{table*}%

\begin{table*}[t]
  \centering
  \setlength\tabcolsep{12pt}
  \resizebox{0.9\textwidth}{!}{
    \begin{tabular}{c|ccccc}
    \toprule
    Method & Art   & Clipart & Product & Real-World & Avg. \\
    \midrule
    D-SAM \cite{d-sam} & 58.03  & 44.37  & 69.22  & 71.45  & 60.77  \\
    JiGen \cite{carlucci2019JiGen} & 53.04  & 47.51  & 71.47  & 72.79  & 61.20  \\
    L2A-OT \cite{zhou2020l2a_ot} & 60.60  & 50.10  & 74.80  & 77.00  & 62.60  \\
    RSC \cite{huang2020rsc} & 58.42  & 47.90  & 71.63  & 74.54  & 63.12  \\
    DDAIG \cite{ddaig} & 59.20±0.10 & 52.30±0.30 & 74.60±0.30 & 76.00±0.10 & 65.50  \\
    MixStyle \cite{mixstyle} & 58.70±0.30 & 53.40±0.20 & 74.20±0.10 & 75.90±0.10 & 65.50   \\
    % DANN \cite{sicilia2021dann} & 61.60  & 48.90  & 75.80  & 76.20  & 65.60  \\
    % DANNCE \cite{sicilia2021dann} & 61.60  & 50.20  & 75.60  & 75.90  & 65.80  \\
    ERM \cite{gulrajani2020search_erm} & 62.70±1.10 & 53.40±0.60 & 76.50±0.40 & 77.30±0.30 & 67.50  \\
    SWAD \cite{cha2021swad}&  66.10±0.40  &  \underline{57.70±0.40}  & 78.40±0.10  &   80.20±0.20 & 70.60 \\
    \midrule
    Ours  & \underline{69.39±0.07} & {57.15±0.17} & \textbf{80.20±0.19} & \textbf{81.79±0.07} & \underline{72.13} \\
    Ours w/ MixStyle & \textbf{69.43±0.19} & \textbf{59.40±0.36} & \underline{78.56±0.16} & \underline{81.85±0.11} & \textbf{72.31} \\
    \bottomrule
    \end{tabular}%
  }
  \vspace{-0.2cm}
  \caption{Results on Office-Home dataset. The best and second-best results are \textbf{bold} and \underline{underlined}, respectively.} %$^*$: The results are taken from the original literature. }
  \vspace{-0.2cm}
  \label{tab:office}%
\end{table*}%

\begin{table*}[!t]
  \centering
  \setlength\tabcolsep{5pt}
  \resizebox{0.9\textwidth}{!}{
    \begin{tabular}{c|ccccccc}
    \toprule
    Method & Clipart & Infograph & Painting & Quickdraw & Real  & Sketch & Avg. \\
    \midrule
    C-DANN \cite{c-dann}$^+$ & 54.60±0.40&17.30±0.10&43.70±0.90&12.10±0.70&56.20±0.40&45.90±0.50&38.30\\
    RSC \cite{huang2020rsc}$^+$ & 55.00±1.20&18.30±0.50&44.40±0.60&12.20±0.20&55.70±0.70&47.80±0.90&38.90\\
    Mixup \cite{mixup}$^+$ & 55.70±0.30&18.50±0.50&44.30±0.50&12.50±0.40&55.80±0.30&48.20±0.50&39.20\\
    SagNet \cite{SagNet}$^+$ & 57.70±0.30&19.00±0.20&45.30±0.30&12.70±0.50&58.10±0.50&48.80±0.20&40.30\\
    MLDG \cite{li2018metalearningDG}$^+$& 59.10±0.20&19.10±0.30&45.80±0.70&13.40±0.30&59.60±0.20&50.20±0.40&41.20\\
    ERM \cite{gulrajani2020search_erm}$^+$& 58.10±0.30&18.80±0.30&46.70±0.30&12.20±0.40&59.60±0.10&49.8±0.40&40.90\\
    MetaReg \cite{balaji2018metareg} & 59.77  & \textbf{25.58 } & 50.19  & 11.52  & 64.56  & 50.09 & 43.62  \\
    DMG \cite{chattopadhyay2020dmg} & 65.24  & 22.15  & 50.03  & 15.68  & 59.63  & 49.02 & 43.63  \\
    SelfReg \cite{kim2021selfreg} & 62.40±0.10 & 22.60±0.10 & 51.80±0.10 & {14.30±0.10} & 62.50±0.20 & {53.80±0.30} & 44.60\\
    % MixStyle \cite{mixstyle}$\ddagger$ & 66.09±0.12 & 23.19±0.18 & 52.30±0.16 & \underline{15.86±0.09} & 64.87±0.12 & 53.12±0.15 & 45.90  \\
    SWAD \cite{cha2021swad}& 66.00±0.10  & 22.40±0.30 & 53.50±0:10 & \underline{16.10±0.20} & 65.80±0.40 & \textbf{55.50±0.30} & 46.50 \\
    \midrule
    Ours  & \textbf{68.07±0.08} & \underline{25.02±0.18} & \underline{53.54±0.37} & 13.74±0.10 & \textbf{67.39±0.02} & {53.59±0.06} & \underline{46.89} \\
    Ours w/ MixStyle & \underline{67.39±0.05} & 24.21±0.08 & \textbf{53.59±0.32} & \textbf{16.21±0.05} & \underline{65.85±0.12} & \underline{}{54.34±0.27} & \textbf{46.93 } \\
    \bottomrule
    \end{tabular}%
  }
  \vspace{-0.2cm}
  \caption{Results on DomainNet dataset. The best and second-best results are \textbf{bold} and \underline{underlined}, respectively.  $^+$: The results are re-implemented and reported by ERM~\protect\cite{gulrajani2020search_erm}} %$^*$: The results are taken from the original literature.
  \vspace{-0.2cm}
  \label{tab:domainnet}%
\end{table*}%

We compare our method with the state-of-the-art methods on PACS, Office-Home and DomainNet datasets, where the results are reported in Table \ref{tab:pacs}, Table \ref{tab:office} and Table \ref{tab:domainnet}, respectively. From these tables, we can observe that: 
\begin{enumerate}[1)]
    \item Our method achieves an average accuracy of \textbf{86.21}, \textbf{72.13} and \textbf{46.89} percent on PACS, Office-Home and DomainNet benchmark datasets, respectively. Compared to other DG methods, our method achieves the state-of-the-art results on most of DG benchmarks, which shows the superiority of our DDG. 
    % Besides, our method do not rely on the exact domain labels, which is more suitable to real-world datasets than those domain-label-dependent DG methods.  
    \item Our method is flexible to combine with other methods to further improve the performance. Taking Mixstyle~\protect\cite{mixstyle} as an example, the performance of our model can be further improved if we insert MixStyle layers into our model. The results can reach \textbf{87.87}, \textbf{72.31} and \textbf{46.93} percent on PACS, Office-Home and DomainNet benchmarks, respectively. It shows that our method is rather robust even on the very strong baselines, i.e. MixStyle.
\end{enumerate}

\subsection{Ablation Studies}
We conduct ablation studies on PACS dataset to evaluate the effectiveness of asymmetric kernel templates and DomainMix respectively, where the results are shown in Table \ref{tab:ablation}.
\begin{table*}[t]
  \centering
  \setlength\tabcolsep{12pt}
  \resizebox{0.9\textwidth}{!}{
    \begin{tabular}{c|cccccc}
    \toprule
    Method & Art Painting   & Cartoon & Photo & Sketch & Avg.  & \#Params \\
    \midrule
    ResNet \cite{resnet} & 84.99±0.25 & 77.64±0.42 & 97.54±0.19 & 71.63±0.41 & 82.95 & 22.43M \\
    SENet \cite{senet} & 82.08±0.36 & 80.70±0.50 & 96.89±0.21 & 75.97±1.84 & 83.91 & 24.83M \\
    Ours w/ 1$\times$ 1 kernel templates & 81.95±0.96 & 78.20±1.86 & 97.50±0.31 & 78.45±0.34 & 84.03 & 28.17M \\
    Ours w/ 3$\times$ 3 kernel templates & 81.10±0.66 & 78.32±0.71 & \textbf{97.64±0.27} & 77.70±1.90 & 83.69 & 66.56M \\
    \midrule
    Ours & \textbf{85.92±0.17} & 79.68±0.42 & 96.65±0.15 & \textbf{82.62±0.27} & \textbf{86.21} & 31.81M \\
    Ours w/o DominMix & 81.18±0.33 & \textbf{80.18±0.73} & 97.35±0.10 & 78.78±1.11 & 84.37 & 31.81M \\
    \bottomrule
    \end{tabular}%
  }
  \vspace{-0.2cm}
  \caption{Results of ablation studies in terms of asymmetric kernel templates and DomainMix on PACS dataset.}
  \label{tab:ablation}%
%   \vspace{-0.2cm}
\end{table*}%

% First, our method outperforms SE\_ResNet50~\protect\cite{senet}. Second, by replacing four different branches in our model with one of the same branches, such as $1\times 1$ or $3\times 3$, the performance drops, which proves that our different channels really have an effect. Third, our method also suffers from performance degradation when no DomainMix training strategy, which proves that the DomainMix can improve dynamic network performance. Please refer to the main text for details.

\paragraph{Effectiveness of Asymmetric Kernel Templates.}
We study the effectiveness of asymmetric design of kernel templates, by employing common kernel templates that consists of four identical kernels with spatial dimensions of $1\times 1$ or $3\times 3$. Besides, we also compare with a channel-wise attention method, namely SENet~\cite{senet}, which is also a popular and powerful architecture in CNN community.
% To some extent, SENet is considered similar to our method when the spatial size of kernel templates are $1\times 1$. 
As shown in Table \ref{tab:ablation}, we can see that: 
\begin{enumerate}[1)]
    \item Our method achieves superior performance to the vanilla ResNet and SENet by a large margin, which suggests the effectiveness of meta-adjusters on DG tasks.
    \item The result of asymmetric kernel templates is obviously better than that of four itentical $1\times1$ kernel templates and $3\times3$ kernel templates, which confirms the effectiveness of the asymmetric design of kernel templates. In-depth analysis of asymmetric kernel templates will be introduced in the visualization section.
    % \item By applying all these two fundamental branches as well as $3\times 1$ kernel and $1\times 3$ kernel branches simultaneously to the residual network, Our algorithm achieves a state-of-the-art performance of $86.21\%$ compared with residual network, which outperforms any single branch applied to it.
\end{enumerate}

\paragraph{Effectiveness of Novel Sample Simulation.} 
In order to verify the effectiveness of the novel sample simulation method, we also conduct ablation studies on DomainMix. Instead of using DomainMix, we train the networks with limited source domains. Table \ref{tab:ablation} shows the comparison results, from which we can observe that our method suffers obvious performance degradation when training without DomainMix strategy. This result demonstrates the importance of novel sample simulation in teaching the meta-adjuster to adapt novel samples from agnostic target domains.

% Table generated by Excel2LaTeX from sheet 'FACT'
%\begin{table*}[!]
%  \centering
    % Table generated by Excel2LaTeX from sheet 'FACT'
%    \begin{tabular}{c|ccccc}
%    \toprule
%          & art   & cartoon & photo & sketch & avg \\
%    \midrule
%    FACT\cite{xu2021fact}* & 89.63 & 81.77 & 96.75 & 84.46 & 88.15 \\
%    FACT\cite{xu2021fact} & 89.71±1.06 & 82.65±0.54 & 96.45±0.23 & 84.78±0.42 & 88.4 \\
%    FACT + IAGN & \textbf{90.54±0.28} & \textbf{84.93±0.41} & \textbf{97.56±0.29} & \textbf{87.70±0.86} & \textbf{90.18} \\
%    \bottomrule
%    \end{tabular}%
%
%  \caption{Combine our model with FACT, *: results are reported by original paper.}
%  \label{tab:factwithiagn}%
%\end{table*}%

\subsection{Visualization}

\begin{figure}[t]
     \centering
     \begin{subfigure}[b]{0.3\columnwidth}
         \centering
         \includegraphics[width=\textwidth]{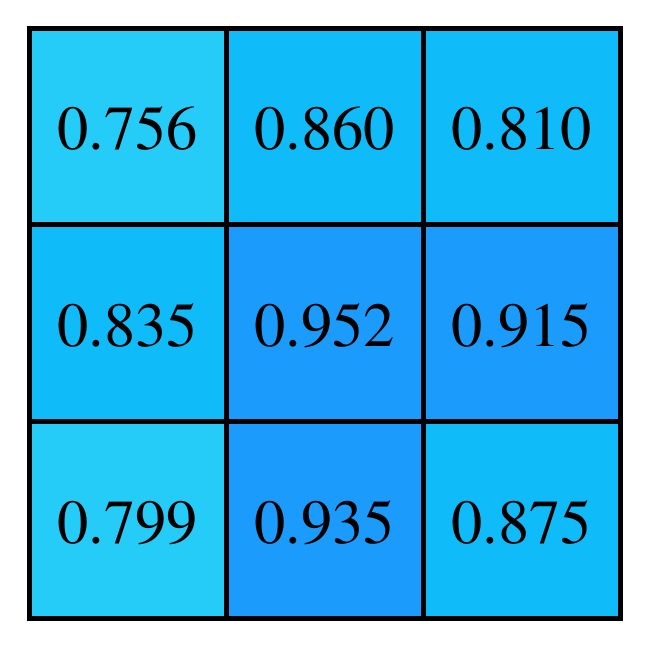}
         \caption{Static kernel}
     \end{subfigure}
     \hspace*{3em}
%     \hfill
%     \begin{subfigure}[b]{0.3\columnwidth}
%         \centering
%         \includegraphics[width=\textwidth]{fig/kernel-visualization-2.pdf}
%         \caption{ACNet}
%     \end{subfigure}
%     \hfill
     \begin{subfigure}[b]{0.3\columnwidth}
         \centering
         \includegraphics[width=\textwidth]{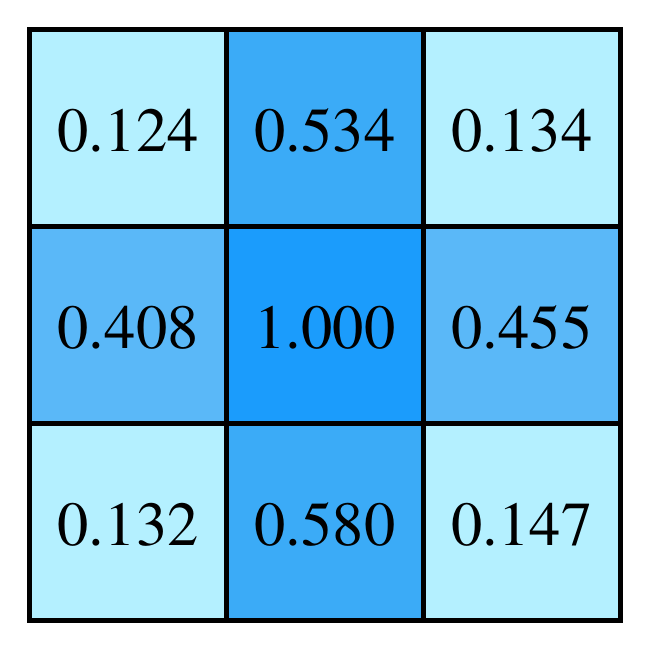}
         \caption{Dynamic kernel}
     \end{subfigure}
        \vspace{-0.2cm}
        \caption{Comparison among static kernel of vanilla ResNet, and dynamic kernel of our network in terms of parameter magnitude. The kernels in sub-figures are acquired by averaging all the kernels, and being normalized in layer dimension.
        % along channels and layers for conventional or dynamic network, respectively. The depth of colors implies the magnitude of parameters.
        }
        \label{fig:kernel_visualize}
        \vspace{-0.2cm}
\end{figure}

\begin{figure}[t]
     \centering
        % \vspace{0.1cm}
     \begin{subfigure}[b]{0.49\columnwidth}
         \centering
         \includegraphics[width=\textwidth]{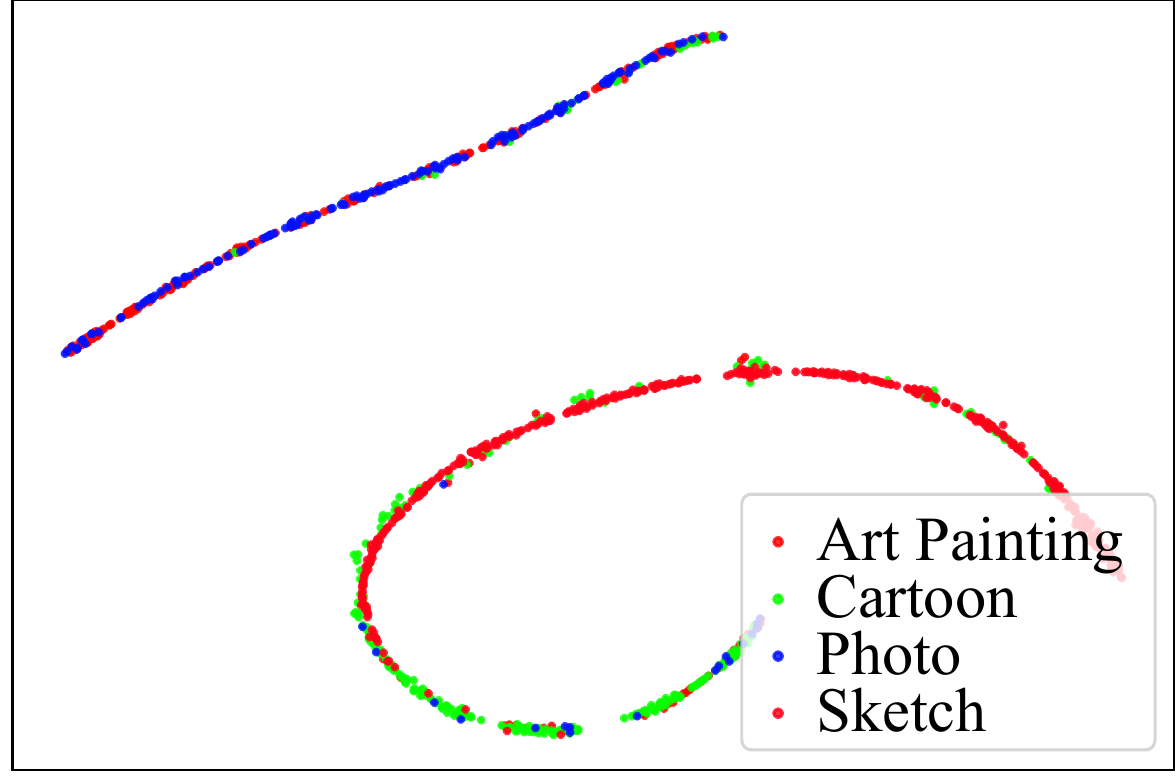}
         \caption{Coefficients of block \#1}
     \end{subfigure}
     \hfill
     \begin{subfigure}[b]{0.49\columnwidth}
         \centering
         \includegraphics[width=\textwidth]{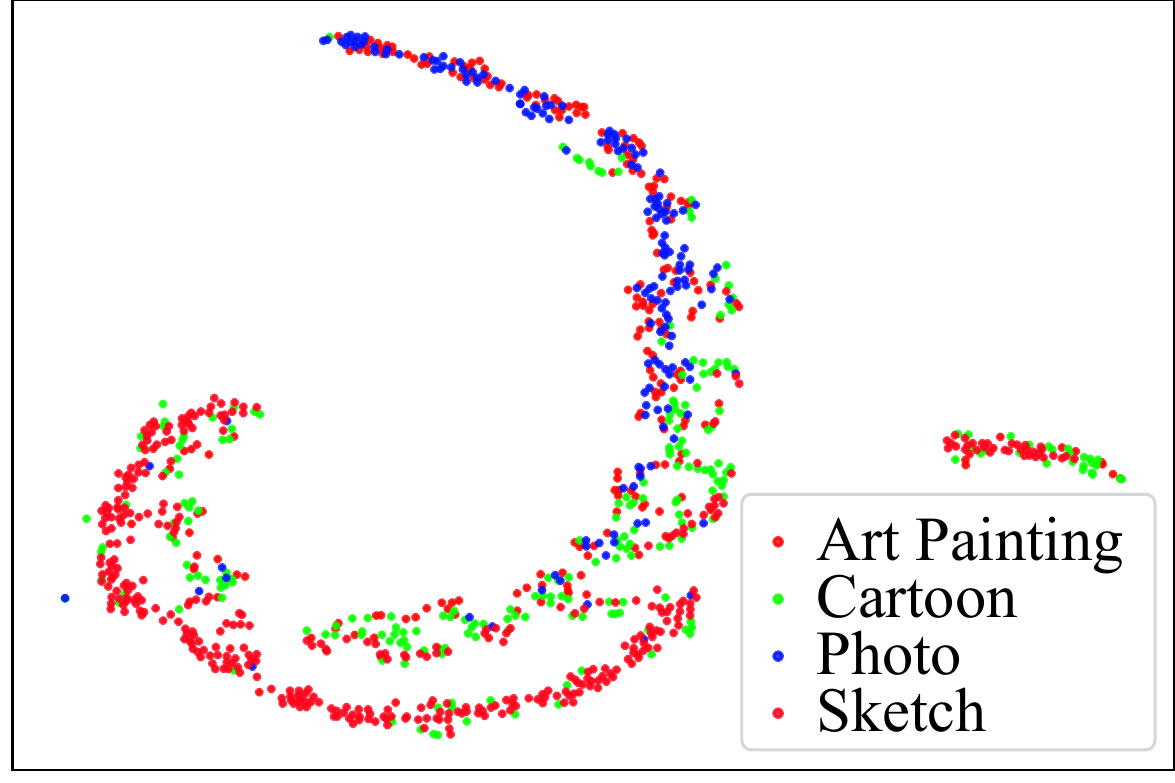}
         \caption{Coefficients of block \#4}
     \end{subfigure}
     \begin{subfigure}[b]{0.49\columnwidth}
         \centering
         \includegraphics[width=\textwidth]{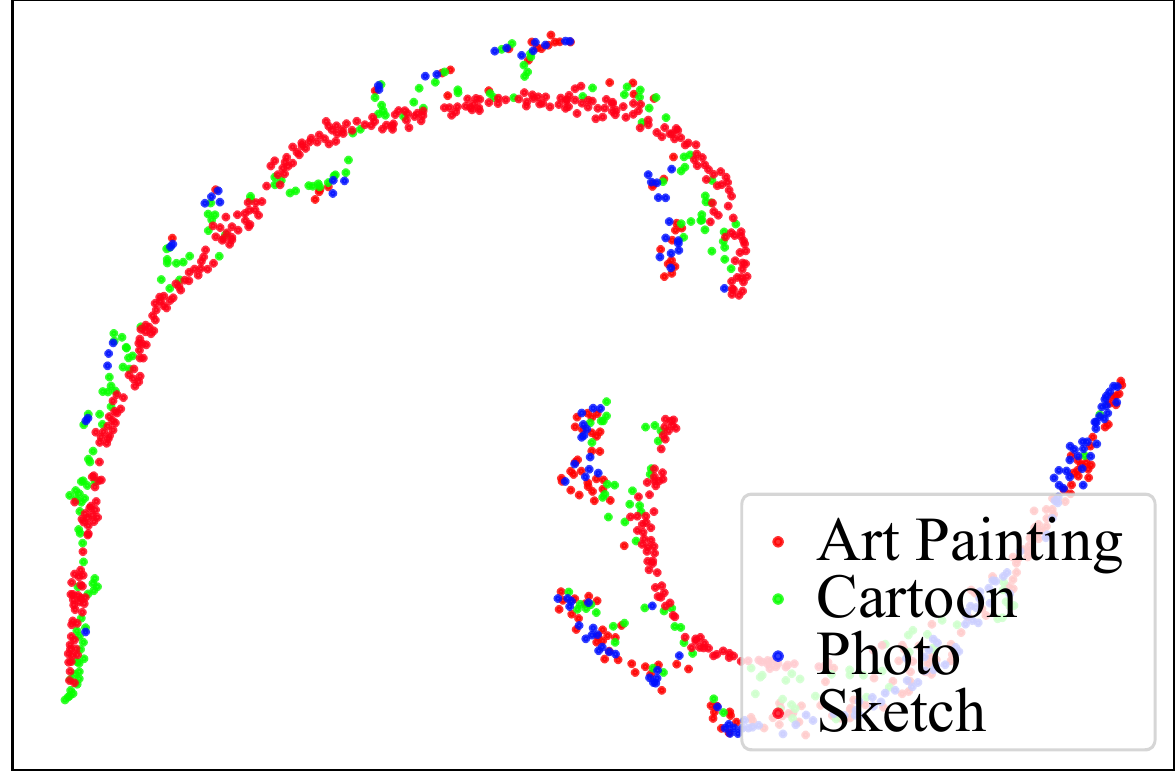}
         \caption{Coefficients of block \#8}
     \end{subfigure}
     \hfill
     \begin{subfigure}[b]{0.49\columnwidth}
         \centering
         \includegraphics[width=\textwidth]{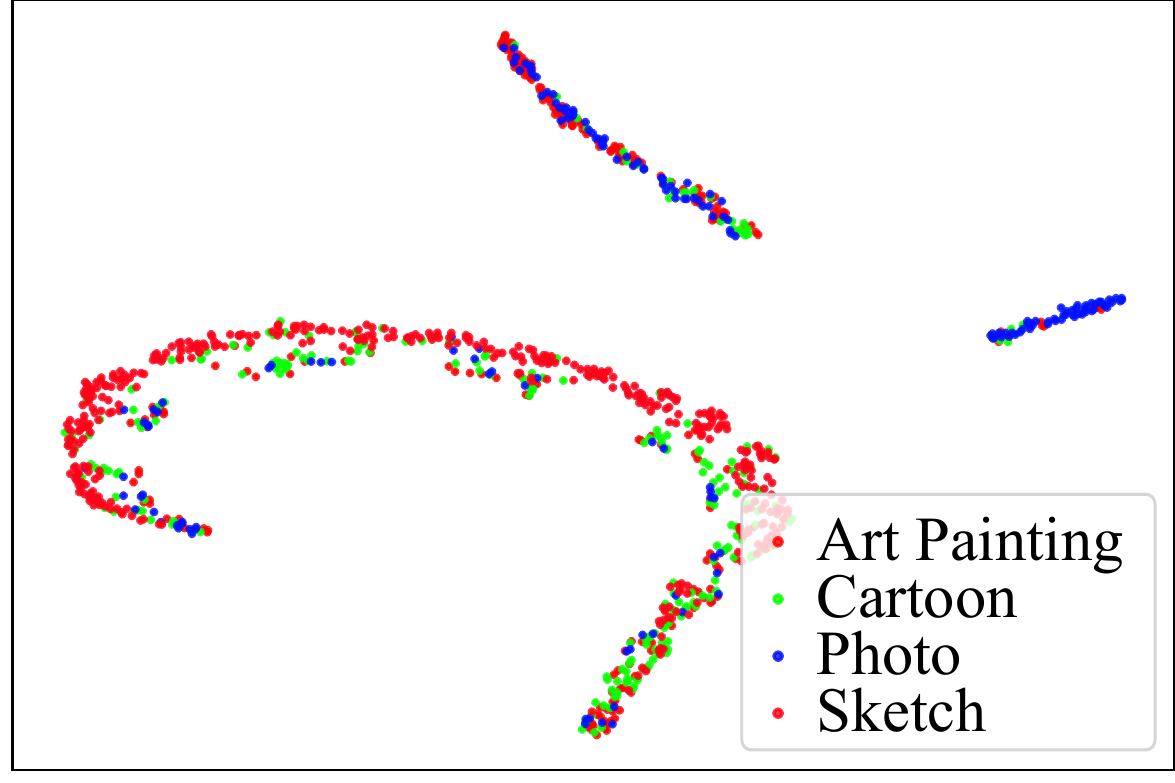}
         \caption{Coefficients of block \#14}
     \end{subfigure}
     \begin{subfigure}[b]{0.49\columnwidth}
         \centering
         \includegraphics[width=\textwidth]{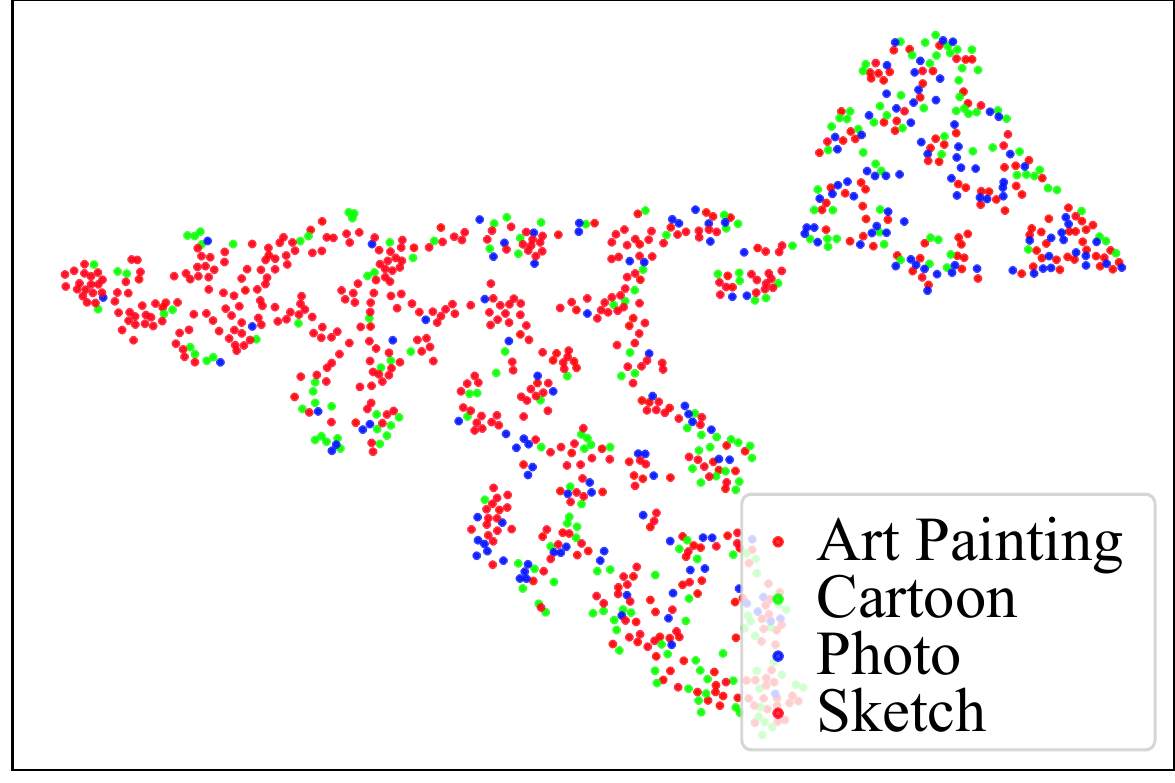}
         \caption{Coefficients of block \#16}
     \end{subfigure}
     \hfill
     \begin{subfigure}[b]{0.49\columnwidth}
         \centering
         \includegraphics[width=\textwidth]{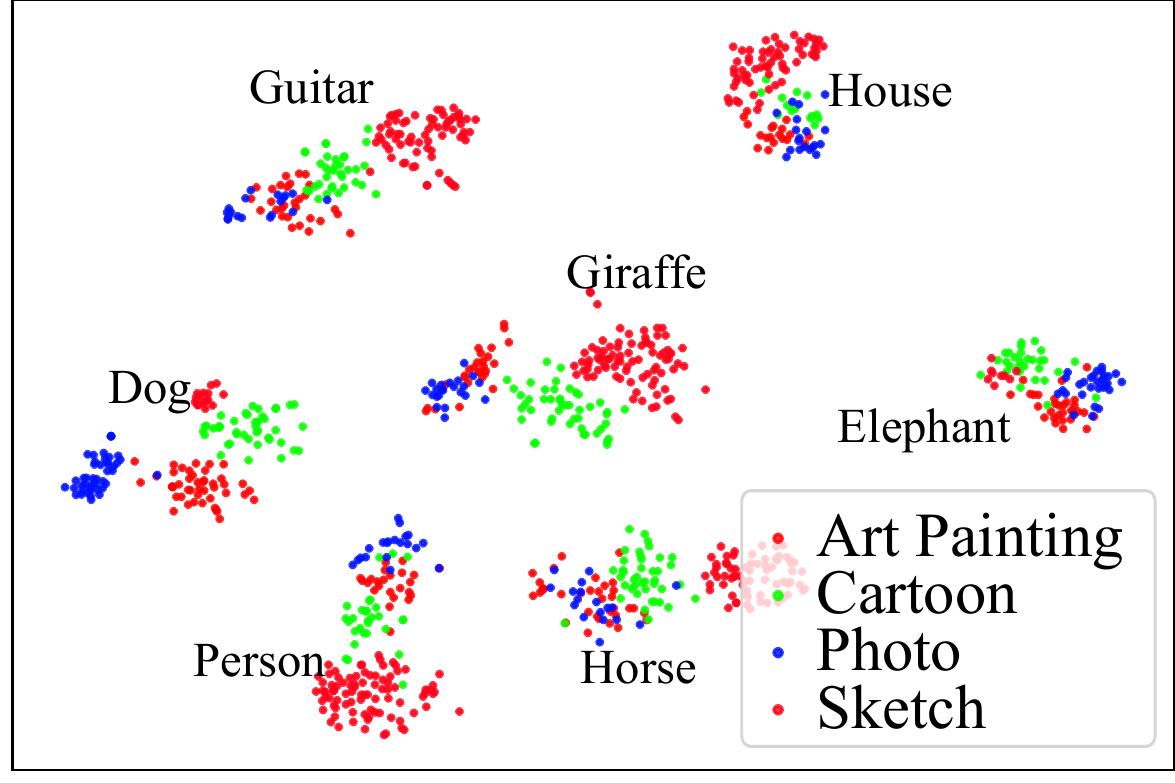}
         \caption{Features}
     \end{subfigure}
     
        % \vspace{-0.1cm}
        \caption{(a)-(e): t-SNE visualization of dynamic coefficients for samples from different domains, where we use colors to distinguish domains. Coefficients of Block \#$a$ denotes the dynamic coefficients from the $a^{th}$ block. (f): t-SNE visualization of features (from the last block) for samples from different domains. Best viewed in colors.}
        \label{fig:tsne}
        % \vspace{-0.1cm}
\end{figure}

For a better understanding of our method, we visualize the kernel templates and dynamic coefficients in different ways. Note that we use the model trained on the multi-source domains (i.e., Art Painting, Cartoon and Sketch) of PACS. 

\paragraph{Visualization of Asymmetric Kernel Templates.}
In order to further explore the mechanism of asymmetric kernel templates, we have a comparison between static kernels in vanilla ResNet and dynamic kernels in our dynamic network, by the means of visualizing \emph{kernel magnitude matrix} following the literature~\cite{ding2019acnet}. For ResNet, we sum up the $3\times 3$ kernels along the channel and layer dimensions to obtain a $3\times 3$ kernel magnitude matrix. For dynamic network, we randomly select an image from validation set to feed into the model and aggregate all the kernel templates with corresponding dynamic coefficients, summing up along the channel and layer dimensions. The comparison result is shown in Figure \ref{fig:kernel_visualize}, from which we can see that:
\begin{enumerate}[1)]
    \item The central criss-cross positions have larger values than the corners of kernel, no matter whether static network or dynamic network. It confirms the kernel skeletons are more crucial than the corners when learning knowledge.  
    \item The kernel skeletons are further strengthened in dynamic network (as shown in Figure \ref{fig:kernel_visualize}(b)). The reason is that the design of dynamic combination among kernel templates can adjust or enlarge the weights of kernel skeletons in instance-aware manner. Consequently, it can promote the model for learning advanced knowledge, which further improves the ability of training-free adapt in inference.  
\end{enumerate}

\paragraph{t-SNE Visualization of Dynamic Coefficients and Features.}
In order to investigate the effect of dynamic network on different distribution data, we select a few blocks to visualize corresponding dynamic coefficients and the features. Figure \ref{fig:tsne}(a)-(e) show the distributions of dynamic coefficients from different blocks of our model, by taking the whole dataset as input, including source (i.e., Art Painting, Cartoon, and Sketch) and target domains (i.e., Photo). We can observe that the dynamic coefficients tend to separate based on domains, and this phenomenon is decelerated from shallow to deep layers. It proves that the instance-aware coefficients are domain-aware practically, even without the explicit supervision of domain labels. Thus, the domain-aware coefficients can realize domain alignment in feature space, as shown in Figure \ref{fig:tsne}(f). It stresses the superiority of dynamic coefficients on learning domain-invariant features for DG tasks.

% Figure \ref{fig:tsne} show the results of t-SNE. For coefficients, ResNet50 has a total of 16 blocks spreading over four residual layers. Limited by the length of the paper, we only select five visualization results from them. For features, we use the output of ResNet backbone for t-SNE visualization. From the Figure \ref{fig:tsne}(a)-\ref{fig:tsne}(e), we can see that for the instances in the same domain, the coefficients are clustered together, while the coefficients of different domain`s instances are clustered in different regions, which shows that our model can aware the domain change of the instances. From the Figure \ref{fig:tsne}(f), we can find that instances from the same class but different domains are clustered together, which shows that our model does have the effect of aligning instances.

\section{Conclusion}
In this paper, we develop a brand-new DG variant termed as dynamic domain generalization, which aims to explore a training-free mechanism to adjust the model to adapt agnostic target domains. To this end, we decouple network parameters into static and dynamic parts to disentangle domain-shared and domain-specific features, where the latter ones are dynamically modulated by meta-adjusters with respect to varied novel samples from different domains. To enable this process, DomainMix and Asymmetric Kernel Templates are utilized to simulate novel samples and impose the meta-adjuster to modulate the models diversely for the samples from a large variety of novel domains. Extensive experiments demonstrate the superiority of our proposed DDG to vanilla DG approaches. Our work provides a strong baselinne for DDG. How to further improve DDG from an optimization perspective is an interesting future work. 
% We hope the proposed DDG can bring new inspirations for DG community.

% We employ DomainMix to simulate data from an uncountable number of domains, which enables the static model to learn domain-shared features and the meta-adjuster to learn domain-specific features, so as to agnostic target domains. We implement our model based on ResNet50~\cite{resnet} and conduct extensive experiments. Results show that our method has excellent generalization performance, outperforming all compared state-of-the-art.

\clearpage

\section*{Acknowledgments}

This work was supported by the National Natural Science Foundation of China (No.: 83321016, 83418022, U21A20471); the Fujian Provincial Youth Education and Scientific Research Project (No. 650722).

%\bibliographystyle{named}
%\bibliography{ijcai22}

\end{document}